\documentclass[letterpaper,10pt,conference]{ieeeconf}
\IEEEoverridecommandlockouts
\usepackage[T1]{fontenc}
\usepackage{graphicx}
\usepackage{booktabs}
\usepackage{amsmath,amssymb,bm}
\let\labelindent\relax
\usepackage{enumitem}
\usepackage{xcolor}
\usepackage{caption}
\usepackage{cuted}
\usepackage{cite}
\usepackage{url}
\usepackage{hyperref}
\hypersetup{hidelinks}

\newcommand{\rose}{\textsc{ROSE}}
\newcommand{\method}{\rose}
\newcommand{\act}{\textsc{ROSE-Act}}

\graphicspath{{figures/}}
\usepackage{caption}
\title{\LARGE \bfseries
ROSE: A Shared Cross-Embodiment Facial-Motion Latent Space for Humanoid Robot Expression}

\author{Yongji Fu$^{1}$, Man Cao$^{2}$, Yi Zhou$^{3}$, Fei Lin$^{4}$, Mark F. Hansen$^{5}$and Guanqun Cao$^{5,\dagger}$
\thanks{$^{1}$Y. Fu is with the Department of Engineering Mathematics, University of Bristol, Bristol, BS16 1QY, United Kingdom.}
\thanks{$^{2}$M. Cao is with the School of Architecture and Design, China University of Mining and Technology, Xuzhou, China.}
\thanks{$^{3}$Y. Zhou is with the Department of Computer Science, Cardiff University, Cardiff, CF10 3AT, United Kingdom.}
\thanks{$^{4}$F. Lin is with Intelliface, Hangzhou, China
 }
\thanks{$^{5}$M. Hansen and G. Cao are with the Bristol Robotics Laboratory, University of the West of England, Bristol, BS16 1QY, United Kingdom.}
\thanks{$^{\dagger}$Corresponding author. Email: \texttt{guanqun.cao@uwe.ac.uk}.}
}

\begin{document}
\maketitle
\thispagestyle{empty}
\pagestyle{empty}
\begin{abstract}
Natural, coherent full-face dynamic expressions are central to making humanoid robots socially approachable, yet existing systems commonly realize only limited lip motion or scripted expressions. The limitation is both physical and algorithmic. A compact head leaves little room for densely packed, independently controllable actuation; meanwhile, learned systems often generate an avatar video, landmarks, or facial parameters and then infer actuator commands through visual retargeting. This render-and-invert path adds latency and artifacts, and it does not reliably transfer generated behavior to a physical robot. We present \method{} (RObot facial expression through Shared motion Embeddings), a framework that integrates biomimetic hardware, cross-embodiment motion representation learning, and multimodal motion generation. \textbf{On the hardware side}, we build a flexible tendon-sheath transmission structure that integrates 34 actuators into a 19.5\,cm biomimetic head. By decoupling actuator placement from facial action points, it enables coordinated full-face expression within a compact volume. \textbf{On the algorithmic side}, ROSE-VAE learns an identity-independent shared facial-motion latent space from human, digital-human, and robot observations. ROSE-Flow directly maps text, audio, or their combination to trajectories in this motion space without using video as a generator input. \act{} then converts these trajectories into constrained actuator commands. Our system demonstrates how high-density biomimetic hardware and a shared executable facial-motion representation can jointly unify cross-embodiment learning, multimodal generation, and physical robot expression.
\end{abstract}

\begin{figure*}[t]
  \centering
  \includegraphics[width=0.9\textwidth]{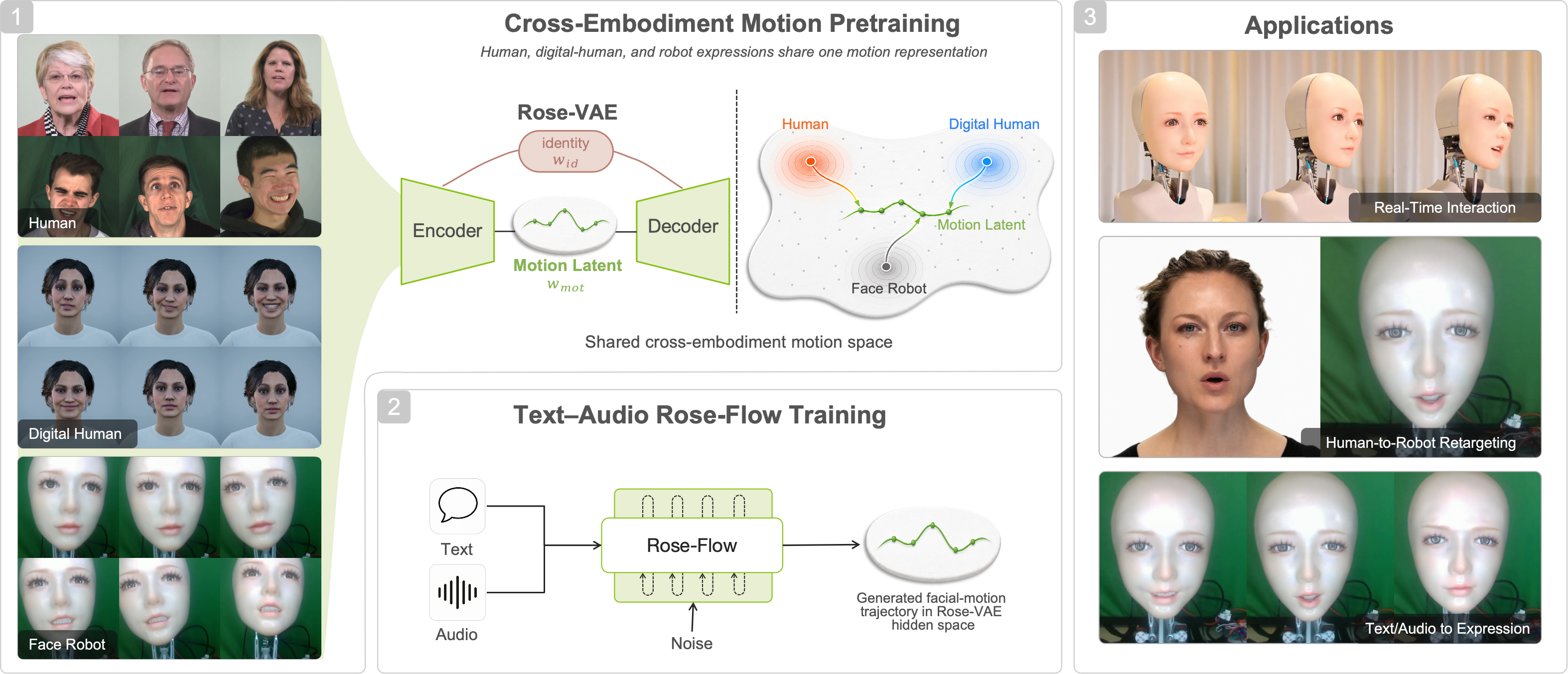}
  \caption{\textbf{ROSE overview.} Human, digital-human, and face-robot observations are mapped by ROSE-VAE into a shared facial-motion latent space. ROSE-Flow generates motion-latent trajectories from text and/or audio, and \act{} maps them to the actuator actions of a real robot. The right side illustrates real-time interaction, human-to-robot retargeting, and text/audio to expression applications. }
  \label{fig:teaser}
  \vspace{-1.0em}
\end{figure*}

\section{Introduction}
\label{sec:intro}
Imagine a humanoid robot turning toward a visitor and answering a question about an exhibit. Its answer may be accurate and its voice natural, yet the interaction can still feel distant if its face remains largely still or moves out of sync with speech. In face-to-face communication, facial movements help people interpret not only what is said but also the speaker's intent and engagement. Coordinating facial expression with speech and context is therefore an important part of making a humanoid robot feel socially present and approachable.~\cite{breazeal2003emotion,bennewitz2005towards,kuno2007museum}.

Reproducing expressive facial motion requires coordinated actuation within a compact head. Previous robotic faces have explored pneumatic and motor-driven mechanisms coupled with compliant skin~\cite{kobayashi2003realization,oh2006design}. These mechanisms provide the physical capacity for expression, but controlling them requires translating desired facial movements into actuator commands compatible with the robot’s mechanical constraints.

Existing approaches often address facial-motion generation and robot execution in separate stages. In the generation stage, audio-driven face animation methods produce visual or parametric outputs, including 3D meshes~\cite{cudeiro2019capture} and portrait videos~\cite{xu2024vasa}. In the execution stage, these outputs are translated into robot actuator commands, for example through visual self-modelling or retargeting of generated facial motion~\cite{hu2024human,hu2026lipmotions}. These image- and parameter-based pipelines benefit from learned facial-motion priors, but connecting the two stages requires converting between representations. Image-based outputs must be analyzed to recover facial motion, while parametric outputs must be mapped to the target robot’s actuator configuration and motion limits. This motivates a shared facial-motion representation that connects human, digital-human, and robot observations with conditional motion generation and actuator control.

We therefore propose \method{} (RObot facial expression through Shared motion Embeddings), illustrated in Fig.~\ref{fig:teaser}. ROSE connects motion encoding, text- and audio-conditioned generation, and robot execution through a shared facial-motion latent space. Generated motion is mapped to constrained actuator commands without intermediate image reconstruction.
Our contributions are:
\begin{enumerate}[leftmargin=*,itemsep=1pt,topsep=2pt]
  \item \textbf{A shared facial-motion representation across embodiments.} ROSE-VAE (variational autoencoder) factorizes identity and motion and learns a shared motion basis with orthogonality regularization across human, digital-human, and robot observations. Its motion coefficients provide a common representation for observation encoding, conditional generation, and actuator mapping.

  \item \textbf{Text- and audio-conditioned motion generation.} ROSE-Flow combines text conditioning with frame-local audio attention in a dual-stream Multimodal Diffusion Transformer (MMDiT). A staged training strategy with model merging unifies audio-to-motion, text-to-motion, and joint text--audio-to-motion generation.

  \item \textbf{Direct latent-to-actuator execution.} \act{} learns to translate shared motion latents into actuator states from paired robot images and actuator states, enabling execution without intermediate image reconstruction. We demonstrate this capability on a compact face robot.

    \item \textbf{A compact biomimetic face robot.} We develop a face robot integrating 34 actuators through a flexible tendon-sheath transmission. By decoupling actuator placement from facial action points, the design supports coordinated full-face expression within a compact head.

\end{enumerate}

\section{Related Work}
\subsection{Biomimetic mechanical faces and expression control.}
Mechanical facial research has long been shaped by two coupled questions: how to place enough independently controllable degrees of freedom in a limited head volume, and how to realize target expressions as stable physical motion. Early platforms used linkages, pneumatics, and compliant skin to realize basic expression and social behavior~\cite{brooks1998cog,kobayashi2003realization,hashimoto2006development,nakaoka2009creating,lin2016expressional,asheber2016humanoid,faraj2021facially}.  
Subsequent work has explored learned facial mimicry, Bayesian optimization, and preference-guided expression generation to achieve more natural facial control~\cite{chen2021smile,yang2022optimizing,tang2023automatic}.
More recent systems incorporate speech, human co-expression, and neural control into physical mechanical faces~\cite{li2024driving,hu2024human,zhang2025morpheus}. Hybrid actuation and visual self-modeling further reveal a trade-off between the precision of rigid mechanisms and the spatial efficiency of tendon transmission~\cite{zhang2025morpheus}. 
Our approach uses a shared facial-motion latent space to connect expression generation with actuator control, allowing generated motion to be mapped to actuator commands without intermediate image reconstruction.

\subsection{Speech- and text-driven facial-motion generation.}
Audio-driven talking-face animation has progressed from visemes and 3D geometric driving~\cite{zhou2018visemenet,cudeiro2019capture} to cross-modal disentanglement, Transformer models, and emotion-conditioned facial motion~\cite{richard2021meshtalk,fan2022faceformer,ji2021audio,peng2023selftalk,peng2023emotalk}, and later to single-image portrait animation and real-time video synthesis~\cite{zhang2023sadtalker,liu2024anitalker,xu2024vasa,guo2024liveportrait,xie2024xportrait}. These methods generally output images, meshes, landmarks, or expression blendshape coefficients and focus on visual appearance, identity preservation, and lip synchronization. Avatar-to-robot retargeting~\cite{hu2026lipmotions} demonstrates the value of large-scale human-face priors, but still depends on an avatar video and a downstream visual bridge. 
In our framework, we instead generate motion-latent trajectories in a shared facial-motion space, where audio provides high-temporal-resolution conditions for articulation, while text specifies global intent and expression style. Video is therefore not an input source at deployment.

\subsection{Disentangled motion representations and conditional flow generation.}
Identity--motion decomposition enables cross-identity animation and reenactment~\cite{wang2022lia,xnemo2025}; in robotics, visual self-modeling also shows that shared visual representations can be connected to embodiment~\cite{chen2022fully}. Diffusion and flow models move generation into continuous latent spaces~\cite{peebles2023dit}, while flow matching and rectified flows offer direct and efficient formulations for conditional trajectory generation~\cite{lipman2023flow,liu2023rectifiedflow}. ROSE combines identity--motion decomposition with conditional flow generation in a shared facial-motion latent space across humans, digital humans, and robots. Frame-local audio conditioning guides motion generation, and a learned actuator mapping connects the generated trajectories to physical execution.

\section{Methodology}
\label{sec:method}
\begin{figure*}[t]
  \centering
  \includegraphics[width=0.8\textwidth]{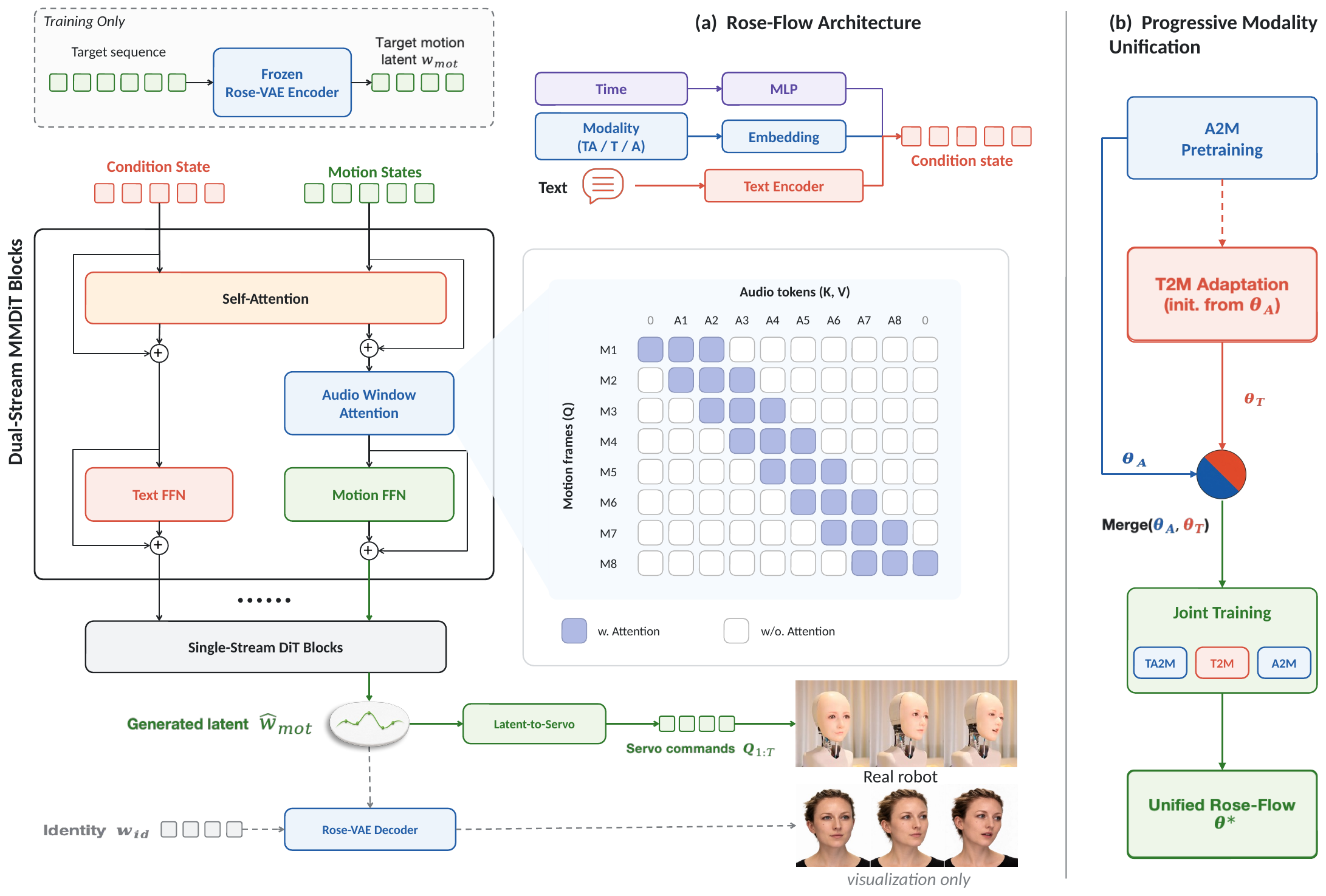}
  \caption{\textbf{ROSE-Flow architecture.} (a)The frozen ROSE-VAE encoder is used during training. The backbone is a text--motion dual-stream MMDiT followed by single-stream DiT blocks that process the concatenated condition and motion tokens. The inset shows the mask of the frame-local audio attention: each motion-frame query attends only to the audio tokens within its local window. Generated motion latents are mapped to real robot actuator commands; decoded faces are shown for visualization only. (b) Progressive modality unification: A2M pretraining, T2M adaptation initialized from $\theta_A$, merging of $\theta_A$ and $\theta_T$, and joint training of the unified model.}
  \label{fig:roseflow}
  \vspace{-1.0em}
\end{figure*}

\subsection{Overview}
Let $I_t$ denote a facial observation at time $t$, which may be a human video frame, a digital-human rendering, or a robot image, and let $q_t\in\mathbb{R}^{J}$ denote the robot actuator state. Given text $c^{\mathrm{text}}$, audio $c^{\mathrm{aud}}$, or both, our goal is to generate an actuator-command trajectory $q_{1:T}$. ROSE connects facial observation, conditional motion generation, and robot execution through a shared motion latent $w^{\mathrm{mot}}_t\in\mathbb{R}^{K}$. ROSE-VAE encodes facial observations $I_t$ into this space, ROSE-Flow generates motion-latent trajectories conditioned on text and audio, and \act{} maps these trajectories to actuator commands. The deployment pipeline is
\begin{equation}
(c^{\mathrm{text}},c^{\mathrm{aud}})
\longrightarrow w^{\mathrm{mot}}_{1:T}
\longrightarrow q_{1:T},
\label{eq:interface}
\end{equation}
where either input modality may be omitted. Generated motion is thus mapped to robot commands without intermediate image reconstruction. We first describe the robot platform, followed by the shared motion representation, conditional generator, and actuator mapping.

\subsection{Robot Platform}
\label{sec:hardware}
\begin{figure}[t]
  \centering
  \includegraphics[width=0.4\textwidth]{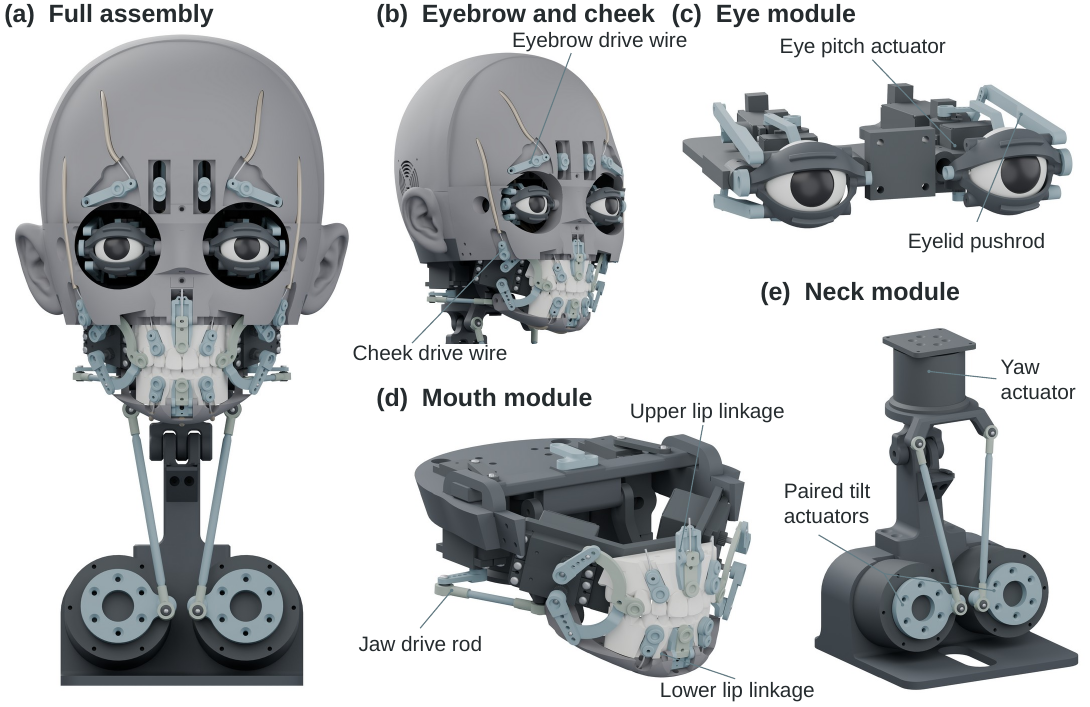}
  \caption{\textbf{Robot platform.} Robot's tendon-sheath transmission and actuator layout.
  }
  \label{fig:robot}
  \vspace{-2.0em}
\end{figure}

Our robot platform is designed to accommodate 34 actuators for facial expression and neck motion. It combines tendon-sheath transmission with local rigid mechanisms to support flexible actuator placement and constrained motion paths. Actuators are concentrated near the top and rear of the head, with tendons routed to the brows, eyelids, cheeks, and lips. Local linkages, pushrods, and shafts additionally drive the jaw, mouth corners, philtrum, and eyeballs, while the neck supports pitch, roll, and yaw. Figure~\ref{fig:robot} shows the prototype and its transmission layout. 

\subsection{ROSE-VAE: Identity--Motion Factorization}
\label{sec:vae}


\noindent\textbf{Explicit identity--motion factorization}
Let $E_{\mathrm{id}}$ and $E_{\mathrm{mot}}$ denote the identity and motion encoders, respectively. Given a reference image $I^{\mathrm{ref}}$ and a target frame $I_t$, ROSE-VAE extracts an identity latent $w^{\mathrm{id}}=E_{\mathrm{id}}(I^{\mathrm{ref}})$ and frame-specific motion coefficients $w^{\mathrm{mot}}_t=E_{\mathrm{mot}}(I_t,I^{\mathrm{ref}})$. The motion coefficients are projected into the latent space through a shared learnable basis $U=[u_1,\ldots,u_K]$ and added to the identity latent:
\begin{equation}
z_t=w^{\mathrm{id}}+U\,w^{\mathrm{mot}}_t.
\label{eq:factorization}
\end{equation}
Here, $U\in\mathbb{R}^{d\times K}$ is a global trainable parameter shared across all samples and domains, where $d$ is the latent dimension and $K$ is the number of motion basis directions. It is initialized with approximately orthogonal columns and jointly optimized with the encoders and decoder through back-propagation. The coefficients $w^{\mathrm{mot}}_t$ determine the contribution of each basis direction, while $U$ defines shared, composable motion directions in the latent space. Orthogonality does not imply that every coordinate represents one semantic facial action; rather, it reduces redundancy and coupling among motion directions and provides a stable interface for later generation and actuator mapping.

\noindent\textbf{Representation Learning Objective}
Human, digital-human, and robot observations share the same encoder, motion basis, and decoder, with facial motion represented by coefficients $w^{\mathrm{mot}}_t$ in a common latent space. For reference $I^{\mathrm{ref}}$ and driving observation $I_t$, the reconstruction is
\begin{align}
\hat I_t = D\!\left(w^{\mathrm{id}}+U\,w^{\mathrm{mot}}_t\right).
\label{eq:vae_reconstruction}
\end{align}
Let $\phi_l(\cdot)$ denote features from layer $l$ of a fixed perceptual network, and let $M_l$ be a binary facial mask at the spatial resolution of perceptual layer $l$. Masks are precomputed from temporally stabilized 2D facial landmarks to cover the mouth and eye regions, then downsampled to each feature scale. 
The training objective is
\begin{equation}
\begin{aligned}
\mathcal{L}_{\mathrm{VAE}}={}&\lambda_{\mathrm{rec}}\mathcal{L}_{\mathrm{rec}}+\lambda_{\mathrm{perc}}\mathcal{L}_{\mathrm{perc}}\\
&+\lambda_{\mathrm{comp}}\mathcal{L}_{\mathrm{comp}}+\lambda_{\mathrm{orth}}\mathcal{L}_{\mathrm{orth}},
\label{eq:vae_loss}
\end{aligned}
\end{equation}

\begin{align}
\mathcal{L}_{\mathrm{rec}}&=\|I_t-\hat I_t\|_1,\label{eq:vae_rec}\\
\mathcal{L}_{\mathrm{orth}}&=\|U^\top U-I\|_F^2,\label{eq:vae_orth}\\
\mathcal{L}_{\mathrm{perc}}&=\sum_l\!\|\phi_l(I_t)-\phi_l(\hat I_t)\|_1,\label{eq:vae_perc}\\
\mathcal{L}_{\mathrm{comp}}&=\sum_l\frac{1}{\lvert M_l\rvert}\left\|M_l\odot\phi_l(I_t)-M_l\odot\phi_l(\hat I_t)\right\|_1.
\label{eq:vae_comp}
\end{align}
Here, $\mathcal{L}_{\mathrm{rec}}$ ensures that recombining identity and motion reconstructs the driven observation; $\mathcal{L}_{\mathrm{perc}}$ preserves global appearance and structure in multi-level perceptual features; $\mathcal{L}_{\mathrm{comp}}$ computes normalized perceptual distances only over masked local facial features, preventing fine expressions in the lips, eyes, and brows from being neglected by global reconstruction~\cite{zhang2018lpips,ki2025float}; and $\mathcal{L}_{\mathrm{orth}}$ keeps the columns of the shared motion basis approximately orthogonal.

\subsection{ROSE-Flow: Text--Audio Conditional Motion Generation}
\label{sec:flow}
ROSE-Flow generates facial-motion latent trajectories in the shared space learned by ROSE-VAE, conditioned on text, audio, or both. Its dual-stream MMDiT processes a current motion-latent sequence together with the available conditions and predicts a velocity field for updating that sequence. Figure~\ref{fig:roseflow} illustrates the overall architecture.

\noindent\textbf{Motion and condition encoding}
Let $x_1=w^{\mathrm{mot}}_{1:T}\in\mathbb{R}^{T\times K}$ be a target motion-latent sequence encoded by the frozen ROSE-VAE encoder, where $T$ is the number of motion frames and $K$ is the motion-latent dimension. During training, we sample Gaussian noise $x_0$ of the same shape and construct the network input by interpolation:
\begin{equation}
x_\tau=(1-\tau)x_0+\tau x_1,\qquad \tau\sim\mathcal{U}[0,1].
\label{eq:flow_path}
\end{equation}
Here, $\tau$ denotes progress from noise to the target sequence, whereas $t$ indexes a motion frame within that sequence. The entire sequence $x_\tau$ serves as the motion states input to the MMDiT. Flow time is encoded as $e_\tau=\mathrm{MLP}(\tau)$.

The conditioning inputs are obtained from text and audio. A text encoder maps the input prompt $p$ to text features $c^{\mathrm{txt}}=E_{\mathrm{txt}}(p)$. For audio, a pretrained Wav2Vec 2.0 encoder extracts acoustic features $F^{\mathrm{aud}}=E_{\mathrm{aud}}(a)$ from waveform $a$. Let $\rho(t)$ denote the acoustic-frame index corresponding to motion frame $t$. With half-window width $h$, we flatten and project the local acoustic features to obtain the audio context for that frame:
\begin{equation}
C_t=\mathrm{Proj}\!\left(\mathrm{vec}\!\left[F^{\mathrm{aud}}_{\rho(t)-h:\rho(t)+h}\right]\right).
\label{eq:audio_pack}
\end{equation}
Here, $\mathrm{vec}$ flattens the local acoustic features into a single vector, and $\mathrm{Proj}$ is a learnable projection that maps it to the required conditioning dimension.
The contexts $C_{1:T}$ collectively form the audio condition $c^{\mathrm{aud}}$ used by ROSE-Flow.

We consider audio-to-motion (A2M), text-to-motion (T2M), and text--audio-to-motion (TA2M). For route $\mu$, the active conditioning inputs are
\begin{equation}
c_\mu=
\begin{cases}
(c^{\mathrm{aud}},e_\mu), & \mu=\mathrm{A2M},\\
(c^{\mathrm{txt}},e_\mu), & \mu=\mathrm{T2M},\\
(c^{\mathrm{txt}},c^{\mathrm{aud}},e_\mu), & \mu=\mathrm{TA2M},
\end{cases}
\label{eq:flow_condition}
\end{equation}
where $e_\mu$ is a learnable embedding identifying the active route. The network therefore receives $x_\tau$, $e_\tau$, and $c_\mu$ at each flow step.

\noindent\textbf{Dual-stream processing and local audio attention}
Within the MMDiT, $x_\tau$ enters the motion stream, while the available text features $c^{\mathrm{txt}}$, flow-time embedding $e_\tau$, and route embedding $e_\mu$ form the condition states. The condition and motion streams interact through joint self-attention. The condition stream then passes through its text feed-forward network, while the motion stream receives a local audio-attention update before its motion feed-forward network. After the dual-stream blocks, the condition and motion tokens are concatenated and refined by a stack of single-stream DiT blocks before the output projection. 

Let $H_t^{\mathrm{mot}}$ denote the hidden features of motion frame $t$ after joint self-attention in the current block. When audio is available, the motion features provide queries, while the audio tokens $c^{\mathrm{aud}}$ provide keys and values:
\begin{equation}
Q_t=H_t^{\mathrm{mot}}W_Q,\
K^{\mathrm{aud}}=c^{\mathrm{aud}}W_K,\
V^{\mathrm{aud}}=c^{\mathrm{aud}}W_V,
\end{equation}
where $W_Q$, $W_K$, and $W_V$ are learnable projection matrices. Each motion frame attends only to audio tokens within its corresponding local window:
\begin{equation}
A_t=\mathrm{softmax}\!\left(\frac{Q_t(K^{\mathrm{aud}})^\top}{\sqrt{d}}+B_t\right)V^{\mathrm{aud}},
\label{eq:local_attn}
\end{equation}
where $d$ is the query/key dimension and $B_t$ masks positions outside the local window by assigning them $-\infty$, with zero entries elsewhere.

The output $A_t$ updates the same motion features through a gated residual connection:
\begin{equation}
\widetilde H_t^{\mathrm{mot}}
=
H_t^{\mathrm{mot}}
+
g\odot\mathrm{Proj}_{\mathrm{out}}(A_t),
\qquad g\big|_{\mathrm{init}}=\bm{0}.
\label{eq:audio_gate}
\end{equation}
Here, $\mathrm{Proj}_{\mathrm{out}}$ maps the attention output to the motion-feature dimension, and $g$ is a learnable gate. The updated features $\widetilde H_t^{\mathrm{mot}}$ pass through the motion feed-forward network and subsequent MMDiT blocks. When audio is absent, this acoustic update is omitted.

For TA2M, text influences the motion features through joint self-attention, while audio supplies frame-local information through the acoustic update. For A2M, the condition states retain $e_\tau$ and $e_\mu$ without text features; for T2M, only the text-conditioning pathway is used. After the final block, the resulting motion features are used to predict a velocity for every frame and latent coordinate. We denote this complete input-to-output mapping by
\begin{equation}
\hat v_\tau=v_\theta(x_\tau,e_\tau,c_\mu)
\in\mathbb{R}^{T\times K},
\label{eq:velocity_prediction}
\end{equation}
where $\theta$ comprises the trainable parameters of the velocity-prediction network.

\noindent\textbf{Flow-matching objective and inference}
The prediction in Eq.~\eqref{eq:velocity_prediction} is trained using the straight path defined in Eq.~\eqref{eq:flow_path}. Its target velocity is
\begin{equation}
\frac{d x_\tau}{d\tau}=x_1-x_0.
\end{equation}
We minimize the squared difference between the predicted and target velocities:
\begin{equation}
\mathcal{L}_{\mathrm{FM}}^\mu
=
\mathbb{E}_{\tau,x_0,x_1}
\left[
\left\|
v_\theta(x_\tau,e_\tau,c_\mu)-(x_1-x_0)
\right\|_2^2
\right].
\label{eq:fm}
\end{equation}

At inference, we initialize $x_0$ with Gaussian noise and repeatedly evaluate the same network under the supplied conditions. Using $N_{\mathrm{step}}$ uniform Euler steps with $\Delta\tau=1/N_{\mathrm{step}}$, we update
\begin{equation}
x_{\tau+\Delta\tau}
=
x_\tau+\Delta\tau\,v_\theta(x_\tau,e_\tau,c_\mu).
\label{eq:flow_inference}
\end{equation}
The updated sequence becomes the motion input for the next flow step. At $\tau=1$, the resulting sequence $\hat w^{\mathrm{mot}}_{1:T}$ is passed to \act{} for actuator-command generation.

\noindent\textbf{Progressive modality unification.} Joint training from scratch can introduce competition between temporally fine-grained audio supervision and abstract text supervision. We therefore first train the A2M model to obtain $\theta_A$, then initialize T2M adaptation from $\theta_A$ by adding text-conditioning modules, yielding $\theta_T$. We interpolate the shared trunk parameters to initialize joint training: \begin{equation} \theta_M=\alpha\theta_A+(1-\alpha)\theta_T,\qquad 0\leq\alpha\leq1, \label{eq:model_merge} \end{equation} while copying the text modules directly from $\theta_T$. Starting from this merged model, we jointly optimize all three routes using a weighted flow-matching objective: \begin{equation} \theta^*=\arg\min_\theta\sum_{\mu\in\{\mathrm{A2M},\mathrm{T2M},\mathrm{TA2M}\}}\omega_\mu\mathcal{L}_{\mathrm{FM}}^\mu. \label{eq:joint_train} \end{equation} Condition dropout and the route identifier in Eq.~\eqref{eq:flow_condition} enable a unified input structure across routes.

\subsection{\act{}: Actuator Mapping}
\label{sec:roseact}

\act{} maps facial-motion latents to robot actuator commands. Through motor babbling, we collect synchronized pairs $(I_t^{\mathrm{rob}},q_t)$, where $I_t^{\mathrm{rob}}$ is a robot image and $q_t$ is the corresponding actuator state. ROSE-VAE extracts the motion coefficients $w^{\mathrm{mot}}_t$ from each robot observation, forming latent--actuator pairs $(w^{\mathrm{mot}}_t,q_t)$ to train the mapping $f_{\mathrm{act}}$.

At inference, \act{} converts the motion-latent trajectory $\hat w^{\mathrm{mot}}_{1:T}$ generated by ROSE-Flow into actuator commands:
\begin{equation}
\hat q_{1:T}=f_{\mathrm{act}}(\hat w^{\mathrm{mot}}_{1:T},s),
\label{eq:servo}
\end{equation}
where $s$ specifies robot calibration, joint limits, and rate limits used to calibrate and constrain the commands before dispatch. The generated motion is thus executed without intermediate image reconstruction.

\section{Experiments and analysis}
\label{sec:experiments}
\subsection{Evaluation Principles and Settings}
We organize experiments around four central questions. \textbf{Q1:} Are motion latents identity-independent and cross-embodiment consistent across humans, digital humans, and face robots? \textbf{Q2:} Can ROSE-Flow simultaneously preserve semantic control, lip synchronization, and natural diversity of facial motion? \textbf{Q3:} How satisfied do users perceive the ROSE-generated robot expressions?


ROSE-VAE is trained on 32{,}021 clips spanning human, digital-human, and face-robot observations. Human-face videos are drawn from public datasets, including HDTF~\cite{zhang2021flow}, RAVDESS~\cite{livingstone2018ryerson}, and VFHQ~\cite{xie2022vfhq}. ROSE-Flow is trained on 24{,}021 clips with text and/or audio supervision. Talking-face videos with synchronized audio, including those from HDTF and RAVDESS, provide supervision for the audio route, while 1{,}760 digital-human clips with expression captions provide supervision for the text route.
\act{} is trained on robot image--actuator-state pairs collected through safe motor babbling, 129{,}431 synchronized frames in total.

\noindent\textbf{Evaluation protocol.} We evaluate talking-face generation on 86 HDTF clips from 78 speakers and 208 RAVDESS clips from two actors. Each clip is truncated to its first 5\,s at 25\,FPS. All methods receive the same first frame and audio. Generated outputs are center-cropped and resized to $512\times512$ and evaluated against the corresponding ground-truth clips.

\noindent\textbf{Evaluation metrics.} 
We measure image and video quality using FID~\cite{heusel2017fid} and 16-frame FVD~\cite{unterthiner2018fvd}. Identity preservation is evaluated using ArcFace cosine similarity, CSIM, to the source frame~\cite{deng2019arcface}. Expression and pose distributions are assessed using E-FID and P-FID computed from Deep3DFaceRecon coefficients~\cite{deng2019accurate}, while lip synchronization is measured using SyncNet LSE-D and LSE-C~\cite{chung2016syncnet}.  For local-expression fidelity, we compare generated and ground-truth frames using mouth LPIPS~\cite{zhang2018lpips} on landmark-aligned mouth crops and blendshape MSE, denoted BS-MSE, over 52 coefficients extracted by a fixed MediaPipe Face Landmarker. On the physical robot, we report retargeting landmark MSE and execution frame rate.

\begin{table}[t]\centering
\caption{\textbf{Cross-identity reenactment for Q1.} 24 driver--target pairs: NeRSemble test sequences of 50 frames each as driving motion and unseen HDTF identities as targets; both motion autoencoders use the same decode path, i.e., target appearance plus driving motion. 
}\label{tab:q1-reenact}
\scriptsize\setlength{\tabcolsep}{3pt}
\begin{tabular}{lcccc}\toprule
Motion autoencoder & CSIM-T $\uparrow$ & CSIM-D $\downarrow$ & AED $\downarrow$ & APD, deg $\downarrow$ \\ \midrule
FLOAT VAE~\cite{ki2025float} & 0.730  & 0.009 & \textbf{0.054} & 1.346  \\
\textbf{ROSE-VAE} & \textbf{0.741 } & \textbf{0.006 }  & 0.264  & \textbf{1.260 } \\
\bottomrule\end{tabular}
\vspace{-1.0em}
\end{table}

\subsection{Identity-Independent Cross-Embodiment Motion Representation (Q1)}


\begin{table*}[t]\centering
\caption{\textbf{Talking-face generation comparison on the HDTF and RAVDESS test lists.} Every method receives the same first frame and audio of up to 5\,s at 25\,FPS; outputs are standardized to 512$\times$512 before scoring. Best in bold, second best underlined.}
\label{tab:talking-face}\small\setlength{\tabcolsep}{5pt}
\resizebox{1.6\columnwidth}{!}{%
\begin{tabular}{llccccccc}\toprule
Dataset & Method & FID $\downarrow$ & FVD $\downarrow$ & CSIM $\uparrow$ & E-FID $\downarrow$ & P-FID $\downarrow$ & LSE-D $\downarrow$ & LSE-C $\uparrow$ \\ \midrule
HDTF, 86 clips & SadTalker~\cite{zhang2023sadtalker} & 60.3 & 193.1 & \underline{0.931} & 0.265 & 0.002 & 10.526 & 5.556 \\
 & AniTalker~\cite{liu2024anitalker} & 31.5 & 109.5 & 0.884 & 0.435 & 0.017 & \underline{10.119} & 6.204 \\
 & Hallo~\cite{xu2024hallo} & 22.2 & 134.9 & 0.904 & 0.565 & 0.004 & 10.219 & \underline{7.201} \\
 & EchoMimic~\cite{chen2024echomimic} & 47.5 & 180.6 & 0.905 & 0.214 & 0.004 & 10.265 & 6.453 \\
 & FLOAT~\cite{ki2025float} & \underline{20.3} & \underline{70.2} & 0.862 & 0.323 & \textbf{0.002} & 10.388 & 7.070 \\
 & \textbf{ROSE, ours} & \textbf{13.5} & \textbf{54.6} & \textbf{0.934} & \textbf{0.167} & \textbf{0.002} & \textbf{9.785} & \textbf{7.773} \\
\midrule
RAVDESS, 208 clips & SadTalker~\cite{zhang2023sadtalker} & 57.9 & 237.8 & 0.921 & 2.471 & 0.022 & 7.450 & 5.150 \\
 & AniTalker~\cite{liu2024anitalker} & 26.0 & 126.5 & 0.880 & \underline{1.499} & 0.018 & 7.773 & \underline{5.497} \\
 & Hallo~\cite{xu2024hallo} & 19.2 & 126.2 & \underline{0.922} & 1.776 & 0.015 & 7.543 & 4.955 \\
 & EchoMimic~\cite{chen2024echomimic} & 39.4 & 182.1 & 0.910 & 2.461 & 0.029 & \underline{7.420} & 4.680 \\
 & FLOAT~\cite{ki2025float} & \underline{15.0} & \underline{81.8} & 0.889 & 1.520 & \underline{0.010} & 7.606 & \textbf{6.003} \\
 & \textbf{ROSE, ours} & \textbf{12.9} & \textbf{56.8} & \textbf{0.924} & \textbf{0.709} & \textbf{0.004} & \textbf{7.298} & 5.445 \\
\bottomrule\end{tabular}}\end{table*}

Table~\ref{tab:q1-reenact} compares ROSE-VAE with the FLOAT motion autoencoder~\cite{ki2025float} on 24 cross-identity reenactment tasks. Each task pairs a 50-frame driving sequence from the NeRSemble test split with an unseen HDTF target identity, testing whether the model can transfer facial motion while retaining the target appearance. ROSE-VAE achieves higher target-identity similarity (CSIM-T: 0.741 vs.\ 0.730) and lower pose error (APD: $1.26^\circ$ vs.\ $1.35^\circ$), indicating better preservation of the target identity and more accurate transfer of head pose. Expression error is comparable (AED: 0.264 vs.\ 0.263), and identity leakage remains low for both models (CSIM-D: 0.009 vs.\ 0.006). Together, these results support the ability of ROSE-VAE to transfer motion across human identities while preserving the target appearance.

\begin{figure}[t]
  \centering
  \setlength{\tabcolsep}{0pt}%
  \renewcommand{\arraystretch}{0}%
  \newcommand{\rowlab}[2]{\rotatebox{90}{\makebox[0.178\columnwidth][c]{\scriptsize\shortstack{#1\\[-1pt]#2}}}}%
  \begin{tabular}{c@{\hspace{2pt}}ccccc}
    & \scriptsize Driving motion & \multicolumn{4}{c}{\scriptsize Target identity} \\[1pt]
    \rowlab{R02}{mouth} &
    \includegraphics[width=0.178\columnwidth]{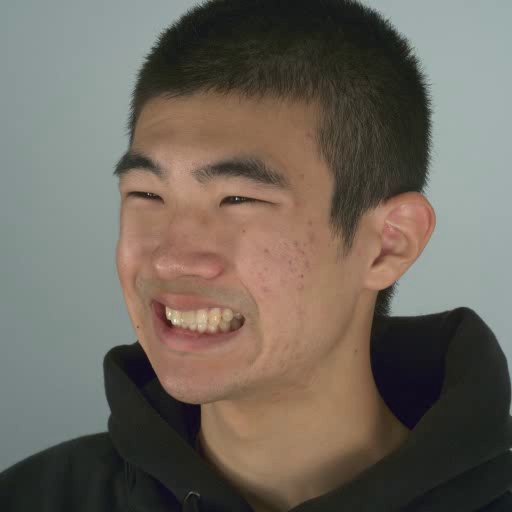} &
    \includegraphics[width=0.178\columnwidth]{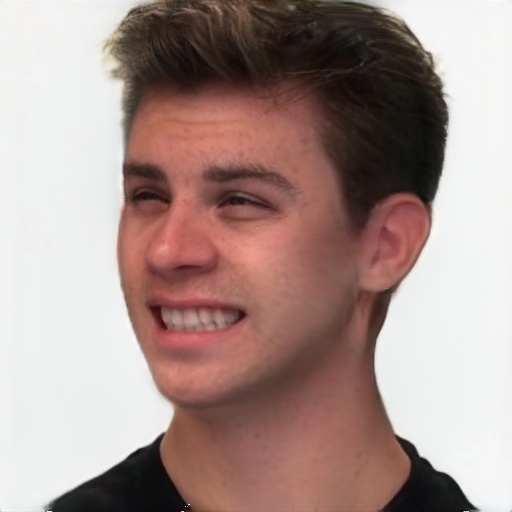} &
    \includegraphics[width=0.178\columnwidth]{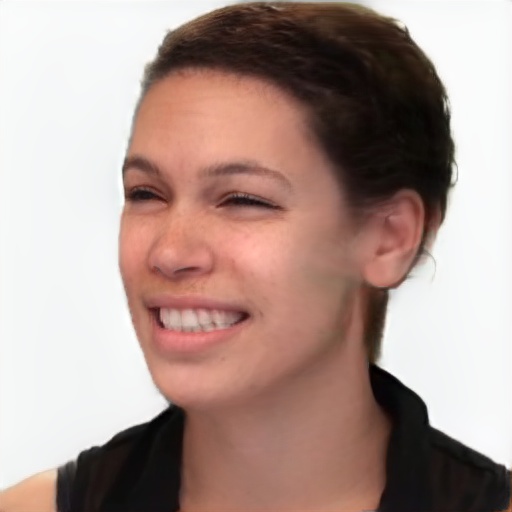} &
    \includegraphics[width=0.178\columnwidth]{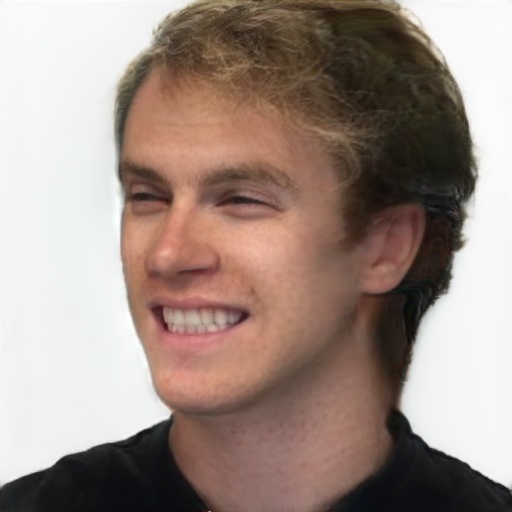} &
    \includegraphics[width=0.178\columnwidth]{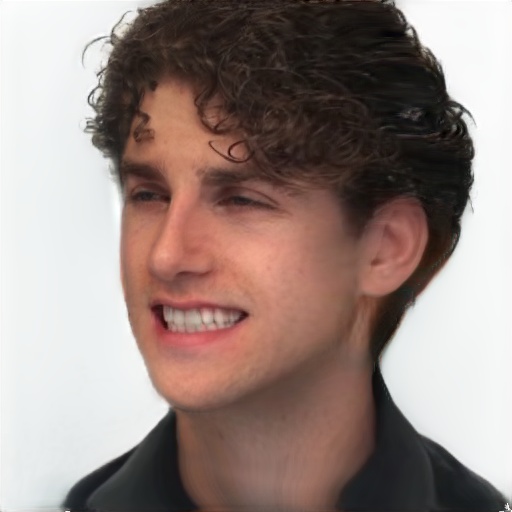} \\
    \rowlab{R04}{surprise, fear} &
    \includegraphics[width=0.178\columnwidth]{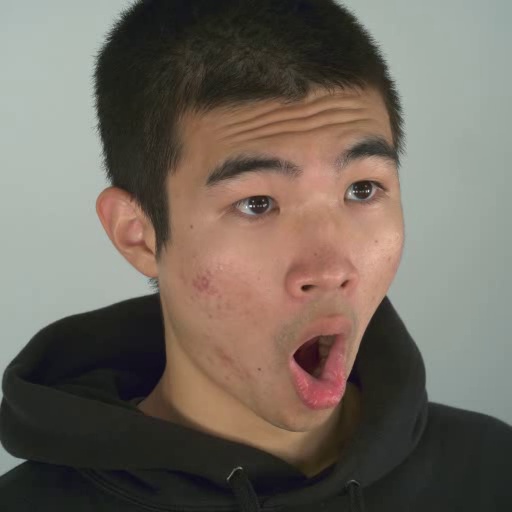} &
    \includegraphics[width=0.178\columnwidth]{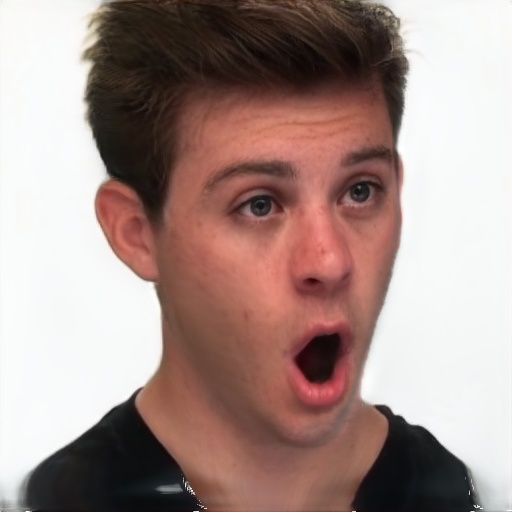} &
    \includegraphics[width=0.178\columnwidth]{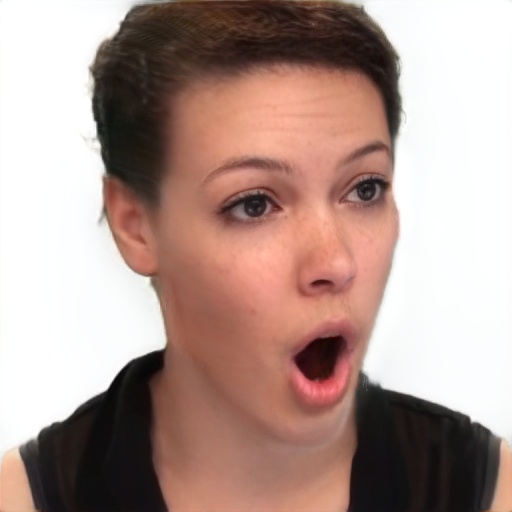} &
    \includegraphics[width=0.178\columnwidth]{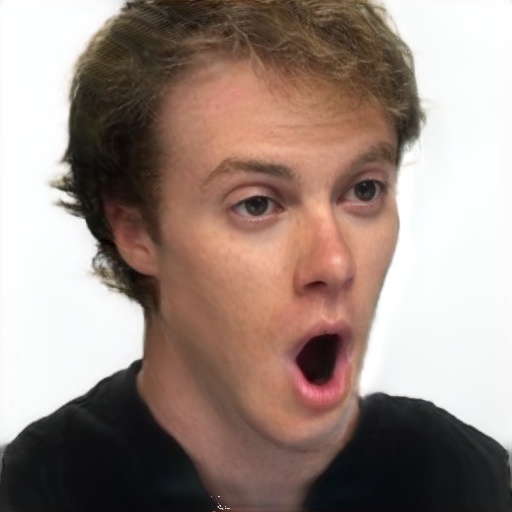} &
    \includegraphics[width=0.178\columnwidth]{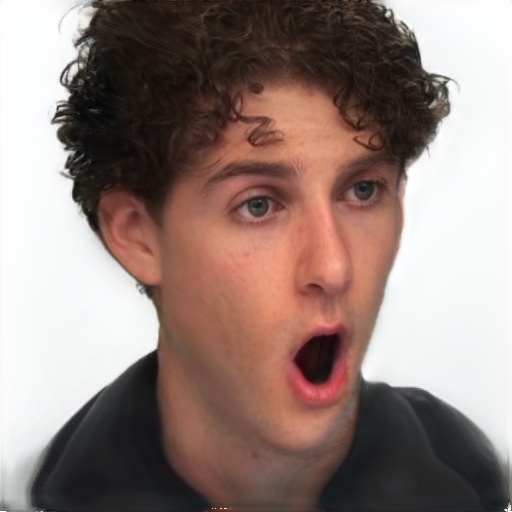} \\
    \rowlab{R24}{surprise, fear} &
    \includegraphics[width=0.178\columnwidth]{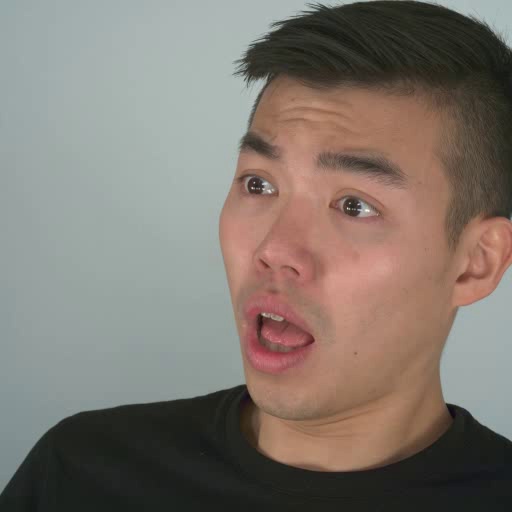} &
    \includegraphics[width=0.178\columnwidth]{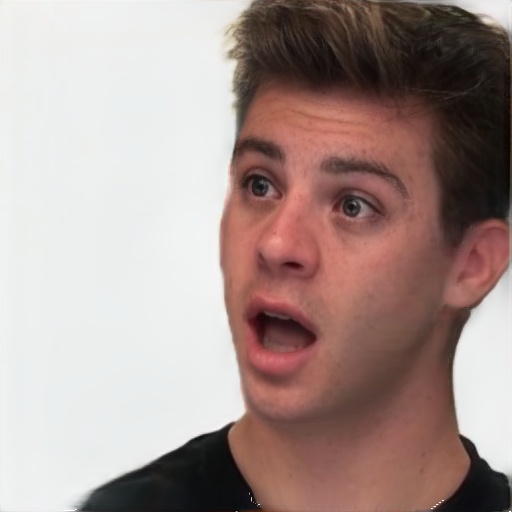} &
    \includegraphics[width=0.178\columnwidth]{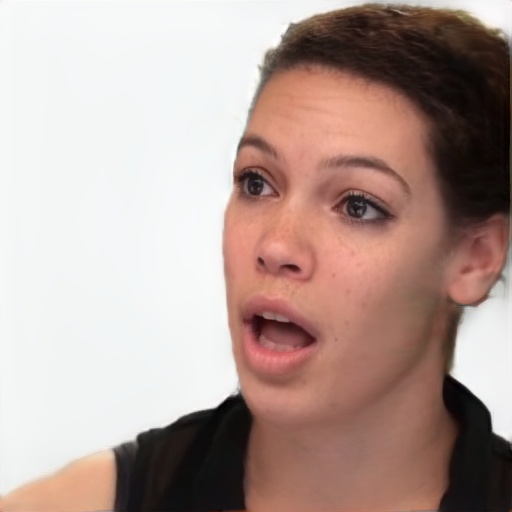} &
    \includegraphics[width=0.178\columnwidth]{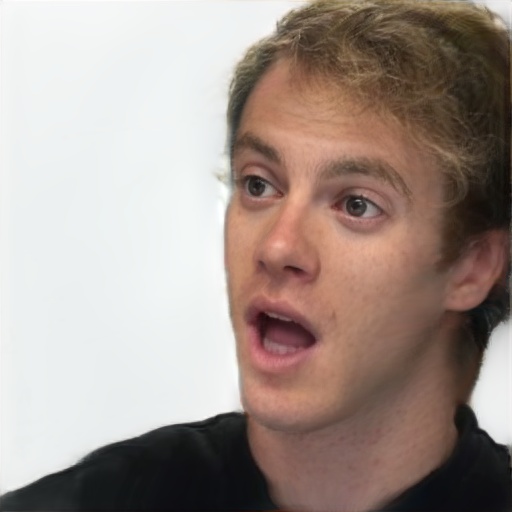} &
    \includegraphics[width=0.178\columnwidth]{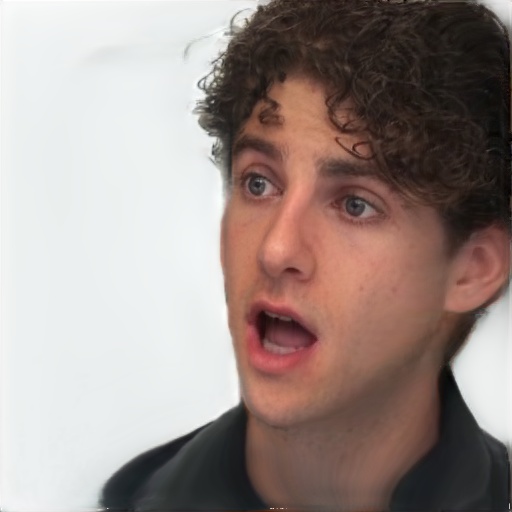}
  \end{tabular}
  \caption{\textbf{Cross-identity motion reenactment.} Each row fixes one driving motion from the NeRSemble test sequences and reenacts it for the same four unseen target identities, one per column.}
  \label{fig:identity-swap}
  \vspace{-2.0em}
\end{figure}

Particularly, we fix three driving motions and combine each with multiple unseen target identities. Figure~\ref{fig:identity-swap} shows all three cross-identity reenactment sets: target appearance and identity features vary with the identity latent, whereas the temporal mouth opening, brow and eye changes, and head pose remain consistent with the driving motion.

\subsection{Semantic Control, Lip Synchronization, and Natural Motion (Q2)}
ROSE-Flow is designed to combine global semantic control from text with local articulatory timing from audio. Text conditions can specify full-face behavior such as smiles, brow raises, surprise, skepticism, gaze, and head pose rather than changing mouth motion alone.
Table~\ref{tab:talking-face} compares ROSE with public talking-face generators, including SadTalker~\cite{zhang2023sadtalker}, AniTalker~\cite{liu2024anitalker}, Hallo~\cite{xu2024hallo}, EchoMimic~\cite{chen2024echomimic}, and FLOAT~\cite{ki2025float}, each evaluated using its official checkpoint and default inference settings. 
ROSE obtains the lowest FID and FVD on both datasets; its FVD of 54.6 on HDTF and 56.8 on RAVDESS compares with 70.2 and 81.8 for the strongest baseline, FLOAT. It also attains the lowest expression and pose distribution distances on RAVDESS, i.e., its videos and their expression and pose statistics stay closest to the recordings. For lip synchronization, ROSE achieves the lowest LSE-D on both datasets, 9.785 on HDTF and 7.298 on RAVDESS, and the highest LSE-C on HDTF, 7.773, demonstrating effective audio--motion alignment alongside the overall fidelity of facial dynamics.



Table~\ref{tab:ablation} reports the available component ablations, and each is evaluated on the HDTF test list of Sec.~\ref{sec:experiments} with the audio-to-motion route, so that every row differs from the full model only in the removed component. It shows that full ROSE achieves the lowest error across all three metrics. Removing the component perceptual loss increases mouth LPIPS from 0.21 to 0.25, while removing local audio attention raises BS-MSE from 0.0164 to 0.0208 and LSE-D from 6.44 to 7.18. Removing progressive modality unification also worsens all three metrics, with LSE-D increasing to 6.89.
\begin{table}[t]
\centering
\caption{\textbf{ROSE component ablations.} Lower is better for all metrics.}
\label{tab:ablation}
\footnotesize
\setlength{\tabcolsep}{2pt}
\resizebox{0.95\columnwidth}{!}{%
\begin{tabular*}{\columnwidth}{@{\extracolsep{\fill}}lccc@{}}
\toprule
Setting & Mouth LPIPS & BS-MSE & LSE-D \\
\midrule
w/o component perceptual loss & 0.25 & 0.0180 & 6.72 \\
w/o local audio attention & 0.24 & 0.0208 & 7.18 \\
w/o progressive modality unification & 0.23 & 0.0193 & 6.89 \\
\textbf{Full ROSE} & \textbf{0.21} & \textbf{0.0164} & \textbf{6.44} \\
\bottomrule
\end{tabular*}%
}
\vspace{-1.0em}
\end{table}

We evaluate physical lip articulation on three RAVDESS~\cite{livingstone2018ryerson} test utterances from one held-out actor, comprising two spoken statements and one sung statement. Table~\ref{tab:lip-articulation-placeholder} compares ROSE with RIT and the NN landmark, Amplitude, and Random baselines.~\cite{hu2026lipmotions}. All methods are implemented on the ROSE robot and evaluated using the same audio inputs. RIT is retrained on our 129{,}431 synchronized robot image--actuator pairs. Following Hu et al., we report frame-wise MSE between the mouth-region latent codes of the recorded robot output and the synthesized reference at 25\,FPS, evaluating lip and jaw articulation.

The results shows that ROSE achieves the lowest mean articulation error on all three test clips. Compared with RIT, ROSE reduces the mean MSE from 0.0202 to 0.0158 on Speech 1, from 0.0169 to 0.0147 on Speech 2, and from 0.0195 to 0.0171 on the song clip. ROSE also yields the lowest maximum error in each clip, indicating smaller peak deviations from the synthesized reference. 

\begin{table}[t]
  \centering
  \caption{\textbf{Controlled lip-articulation comparison.} Mean squared error, MSE, over two spoken and one sung RAVDESS test utterance. All baselines follow Hu et al.~\cite{hu2026lipmotions} and are re-implemented on the ROSE robot, evaluated on the same utterances and protocol.}
  \label{tab:lip-articulation-placeholder}
  \scriptsize
  \setlength{\tabcolsep}{3.2pt}
  \renewcommand{\arraystretch}{1.02}
  \begin{tabular*}{\columnwidth}{@{\extracolsep{\fill}}lcccccc@{}}
    \toprule
    Statistic & \textbf{ROSE} & RIT~\cite{hu2026lipmotions} & NN landmark & Amplitude & Random \\
    \midrule
    \multicolumn{6}{c}{\emph{Speech 1: ``Kids are talking by the door'', 95 frames, 3.8\,s}} \\
    Mean & \textbf{0.0158} & 0.0202 & 0.4520 & 0.6855 & 0.5985 \\
    Std. & \textbf{0.0084} & 0.0131 & 0.3190 & 0.3242 & 0.4491 \\
    Min. & \textbf{0.0031} & 0.0033 & 0.0137 & 0.1827 & 0.0379 \\
    Max. & \textbf{0.0512} & 0.0759 & 1.8102 & 1.7660 & 3.3223 \\
    \addlinespace[2pt]
    \multicolumn{6}{c}{\emph{Speech 2: ``Dogs are sitting by the door'', 92 frames, 3.7\,s}} \\
    Mean & \textbf{0.0147} & 0.0169 & 0.4034 & 0.8295 & 0.5969 \\
    Std. & \textbf{0.0071} & 0.0086 & 0.2723 & 0.2738 & 0.4426 \\
    Min. & \textbf{0.0028} & 0.0038 & 0.0176 & 0.2021 & 0.0332 \\
    Max. & \textbf{0.0512} & 0.0645 & 1.5855 & 2.1023 & 2.6603 \\
    \addlinespace[2pt]
    \multicolumn{6}{c}{\emph{Song: ``Kids are talking by the door'', 110 frames, 4.4\,s}} \\
    Mean & \textbf{0.0171} & 0.0195 & 0.4318 & 0.9016 & 0.6180 \\
    Std. & 0.0093 & \textbf{0.0090} & 0.2852 & 0.2954 & 0.3825 \\
    Min. & \textbf{0.0036} & 0.0050 & 0.0156 & 0.3311 & 0.0407 \\
    Max. & \textbf{0.0588} & 0.0632 & 1.3519 & 1.9127 & 2.2786 \\
    \bottomrule
  \end{tabular*}
  \vspace{-1.0em}
\end{table}

\subsection{Real-Robot Deployment (Q3)}
The shared facial-motion latent space lets ROSE-Flow outputs be converted directly to actuator commands by \act{}, without image generation, landmark extraction, smoothing, and motion retargeting. On the current physical system, this interface increases execution frame rate from 15 FPS to 60 FPS. Real-time interaction, human-to-robot expression retargeting, demonstrating its usability in the physical system. Figure~\ref{fig:app:qualitative-dynamics} shows three representative expression sequences executed on the robot.

Finally, we complement the objective measurements with a user study. Participants rate anonymized clips generated by ROSE and the four baselines in Table~\ref{tab:lip-articulation-placeholder}, all executed on the ROSE robot using the same audio inputs, on five-point scales for realism, emotional resonance, and lip synchronization. As shown in Table~\ref{tab:user-study}, ROSE achieves the highest mean ratings across all three criteria, scoring 3.41, 3.36, and 3.78, respectively, compared with 2.98, 2.73, and 3.40 for RIT.

\begin{table}[t]\centering
\caption{\textbf{User study.} Mean five-point ratings of real-robot clips for realism, emotional resonance, and lip synchronization; baselines as in Table~\ref{tab:lip-articulation-placeholder}, executed on the ROSE robot.}\label{tab:user-study}
\setlength{\tabcolsep}{6pt}
\begin{tabular}{lccc}\toprule
Method & Realism $\uparrow$ & Resonance $\uparrow$ & Lip-Sync $\uparrow$ \\ \midrule
Random selection & 1.34 & 1.41 & 1.38 \\
Amplitude BL & 1.52 & 1.47 & 1.69 \\
NN landmark BL & 1.72 & 1.66 & 2.05 \\
RIT~\cite{hu2026lipmotions} & 2.98 & 2.73 & 3.40 \\
\textbf{ROSE} & \textbf{3.41} & \textbf{3.36} & \textbf{3.78} \\ \bottomrule
\end{tabular}
\vspace{-1.0em}
\end{table}

\section{Conclusion}
\label{sec:conclusion}
We presented \method{}, a shared cross-embodiment facial-motion latent space for text- and audio-driven humanoid robot expression. ROSE-VAE learns identity--motion factorization across human, digital-human, and face-robot observations. ROSE-Flow generates motion-latent trajectories in the same space from text, audio, or both. \act{} maps these trajectories directly to commands for a real robot without first synthesizing robot video. Together with a compact tendon-sheath face robot, ROSE recasts expression perception, generation, and physical control as learning and execution around a common motion variable. Existing measurements support multimodal complementarity, streaming stability, and faster, lower-error physical retargeting.

\begin{figure*}[t]
  \centering
  \resizebox{0.85\textwidth}{!}{%
\begingroup\setlength{\parindent}{0pt}\definecolor{stageaction}{RGB}{180,30,40}
\setlength{\tabcolsep}{1.2pt}\renewcommand{\arraystretch}{0}
\begin{tabular}{@{}ccccccccc@{}}
\multicolumn{3}{c}{\parbox[b]{0.318\textwidth}{\centering\fontsize{7.5}{8.5}\selectfont \textbf{(a)} Slowly opens the mouth as the eyes narrow,}} & \multicolumn{3}{c}{\parbox[b]{0.318\textwidth}{\centering\fontsize{7.5}{8.5}\selectfont then opens wide as if yawning,}} & \multicolumn{3}{c}{\parbox[b]{0.318\textwidth}{\centering\fontsize{7.5}{8.5}\selectfont before closing the mouth and reopening the eyes.}} \\[1.5pt]
\includegraphics[width=0.1035\textwidth]{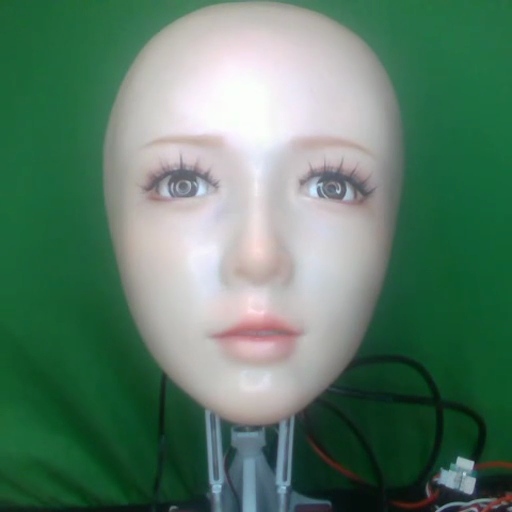} & \includegraphics[width=0.1035\textwidth]{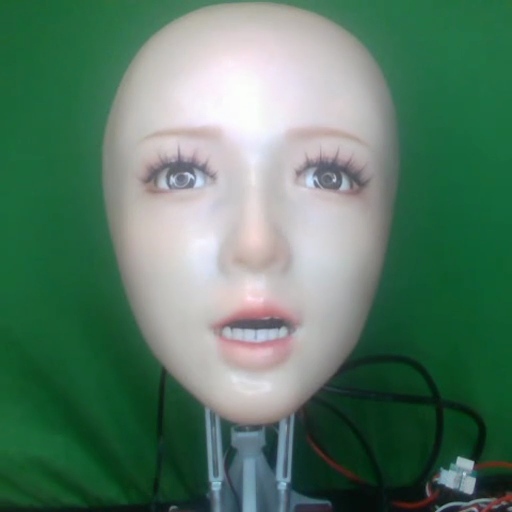} & \includegraphics[width=0.1035\textwidth]{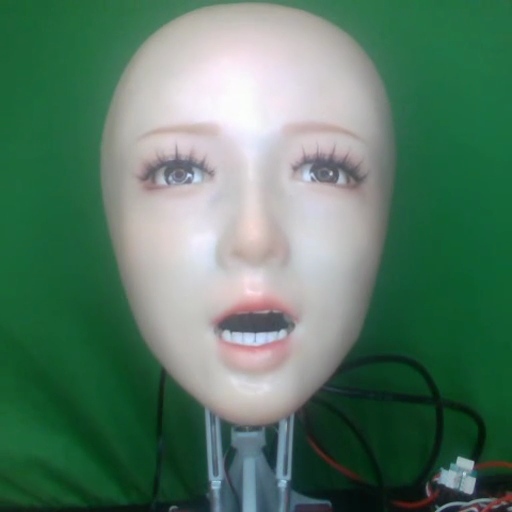} & \includegraphics[width=0.1035\textwidth]{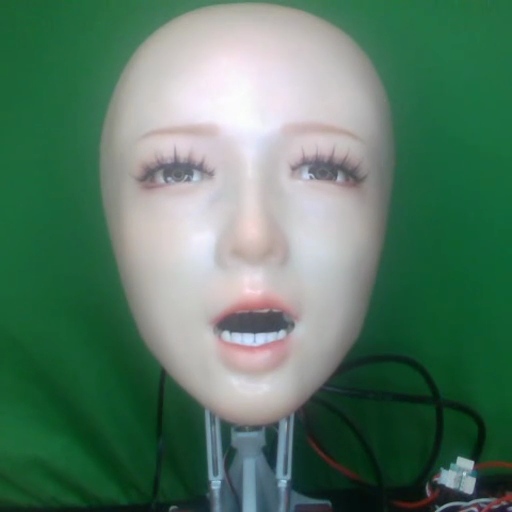} & \includegraphics[width=0.1035\textwidth]{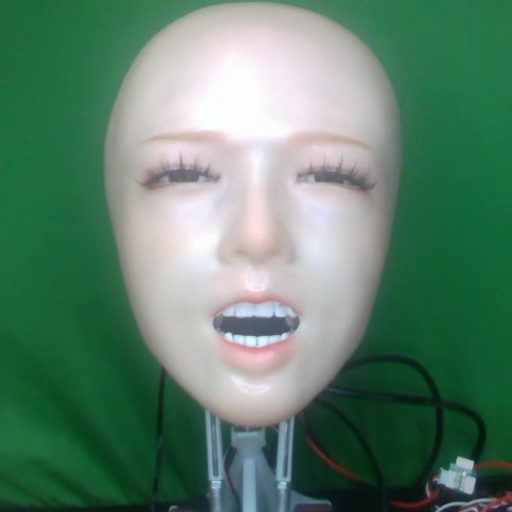} & \includegraphics[width=0.1035\textwidth]{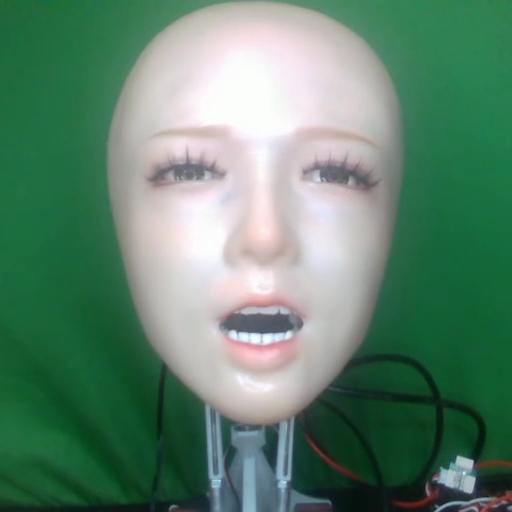} & \includegraphics[width=0.1035\textwidth]{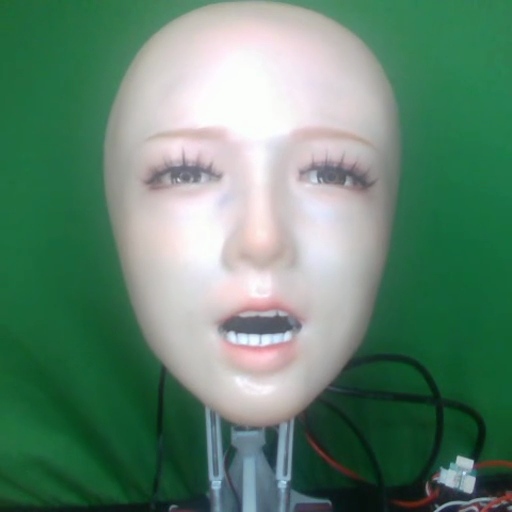} & \includegraphics[width=0.1035\textwidth]{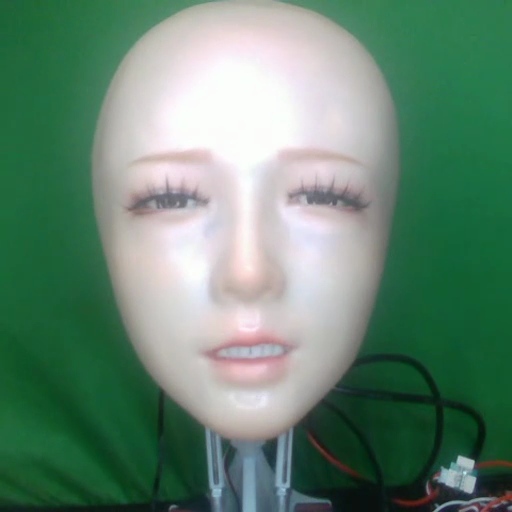} & \includegraphics[width=0.1035\textwidth]{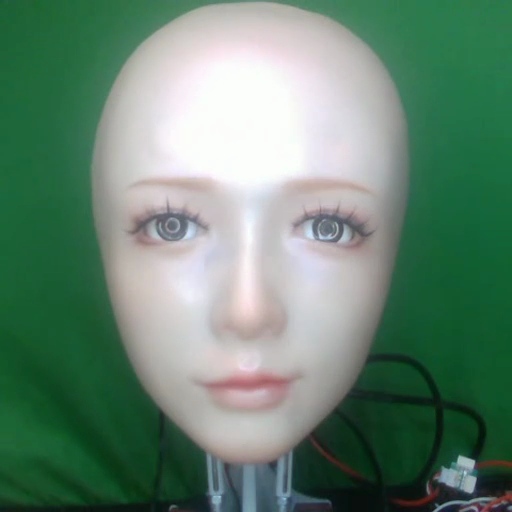} \\[4pt]
\multicolumn{3}{c}{\parbox[b]{0.318\textwidth}{\centering\fontsize{7.5}{8.5}\selectfont \textbf{(b)} Closes the eyes slowly and opens them again,}} & \multicolumn{3}{c}{\parbox[b]{0.318\textwidth}{\centering\fontsize{7.5}{8.5}\selectfont then raises the eyebrows attentively,}} & \multicolumn{3}{c}{\parbox[b]{0.318\textwidth}{\centering\fontsize{7.5}{8.5}\selectfont before gently tilting the head with a faint smile.}} \\[1.5pt]
\includegraphics[width=0.1035\textwidth]{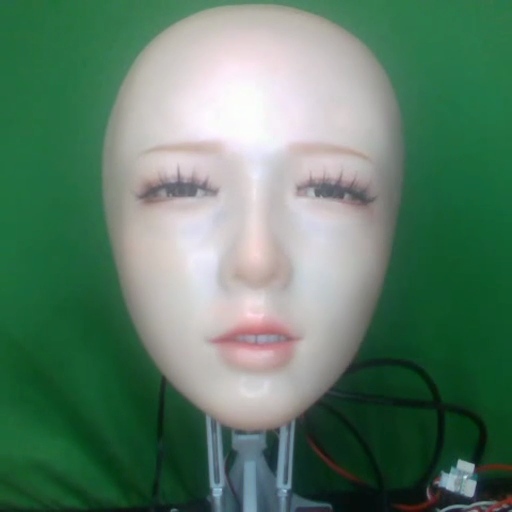} & \includegraphics[width=0.1035\textwidth]{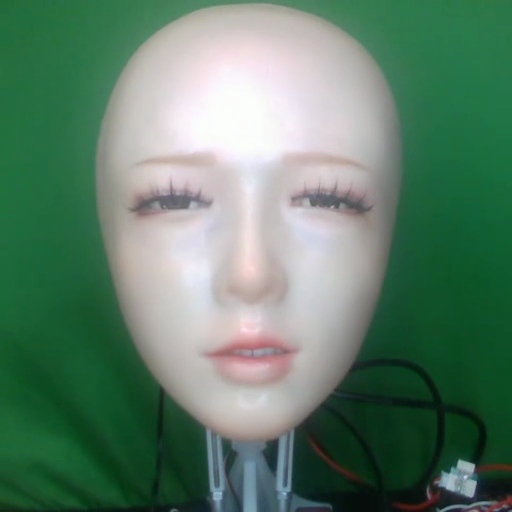} & \includegraphics[width=0.1035\textwidth]{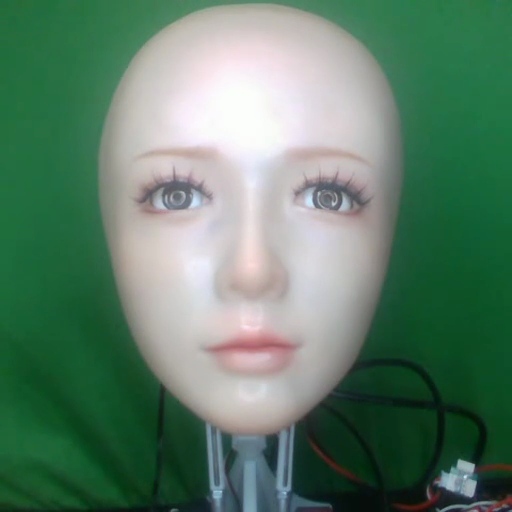} & \includegraphics[width=0.1035\textwidth]{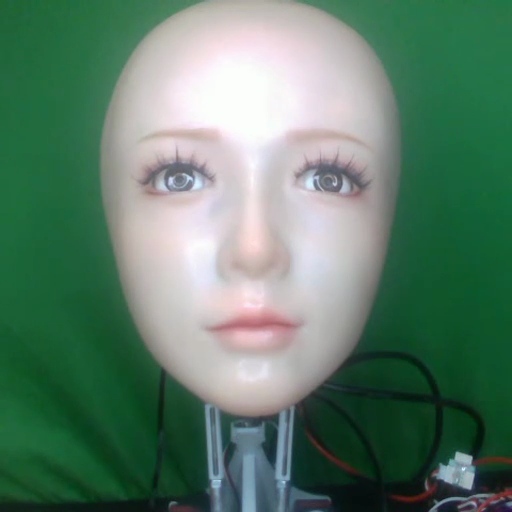} & \includegraphics[width=0.1035\textwidth]{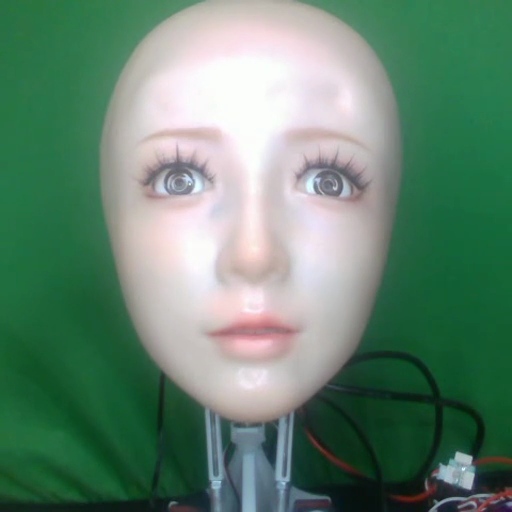} & \includegraphics[width=0.1035\textwidth]{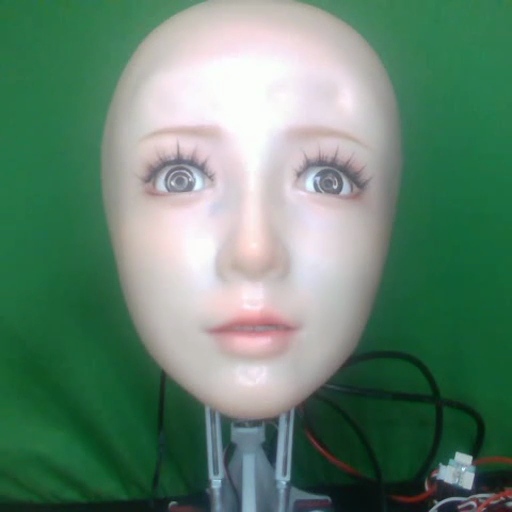} & \includegraphics[width=0.1035\textwidth]{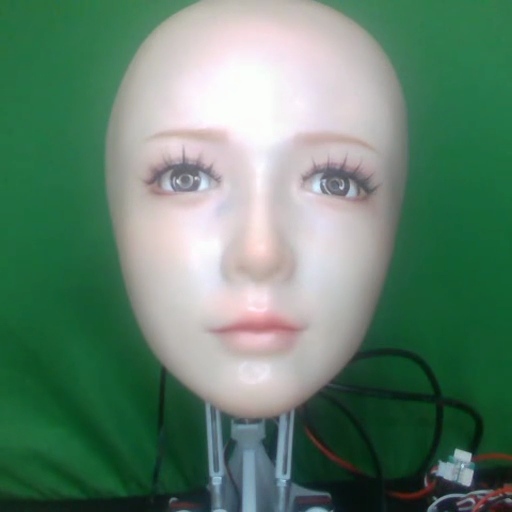} & \includegraphics[width=0.1035\textwidth]{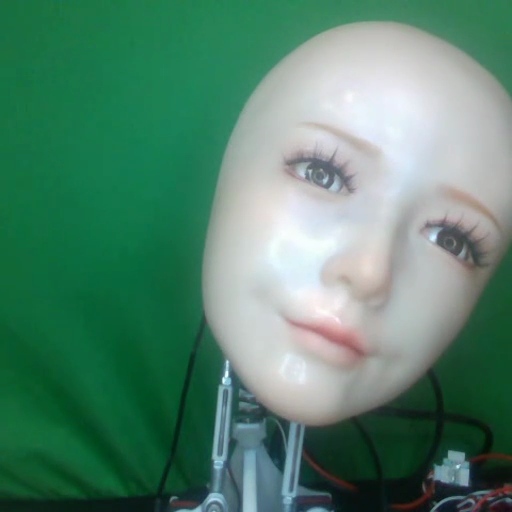} & \includegraphics[width=0.1035\textwidth]{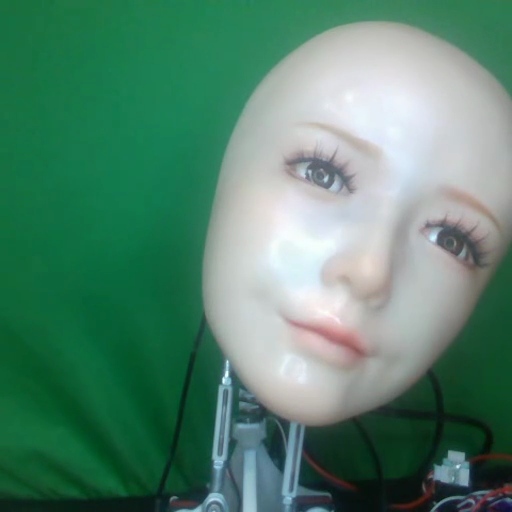} \\[4pt]
\multicolumn{3}{c}{\parbox[b]{0.318\textwidth}{\centering\fontsize{7.5}{8.5}\selectfont \textbf{(c)} Rolls the eyes upward with a weary look,}} & \multicolumn{3}{c}{\parbox[b]{0.318\textwidth}{\centering\fontsize{7.5}{8.5}\selectfont then widens the eyes with an animated expression,}} & \multicolumn{3}{c}{\parbox[b]{0.318\textwidth}{\centering\fontsize{7.5}{8.5}\selectfont before giving a playful blink and a faint smile.}} \\[1.5pt]
\includegraphics[width=0.1035\textwidth]{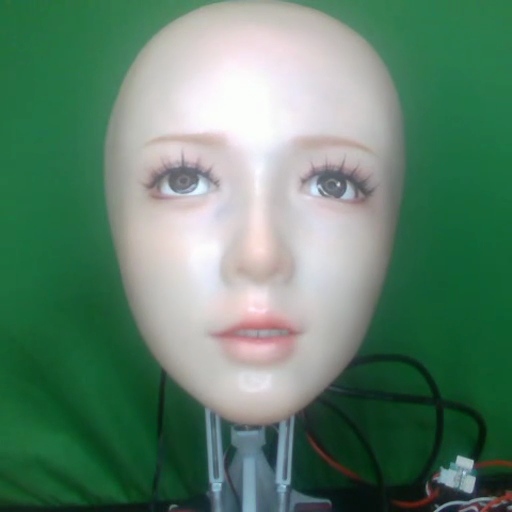} & \includegraphics[width=0.1035\textwidth]{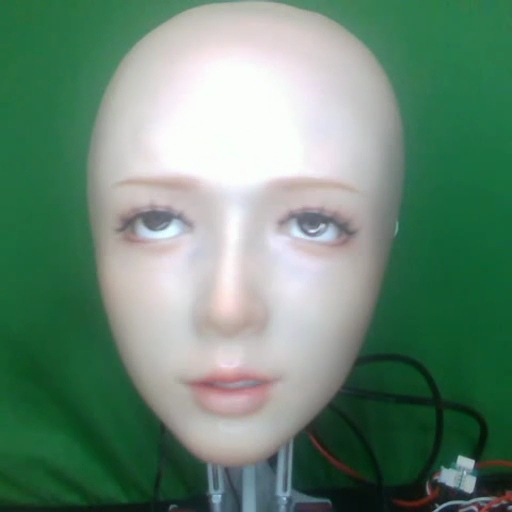} & \includegraphics[width=0.1035\textwidth]{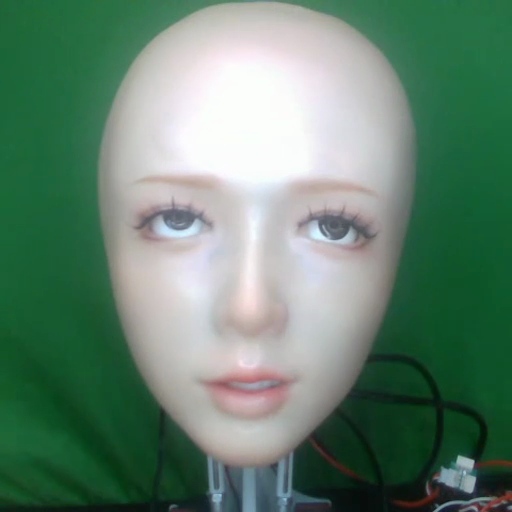} & \includegraphics[width=0.1035\textwidth]{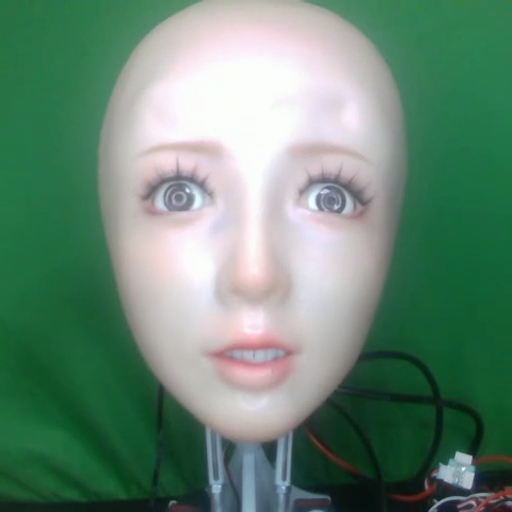} & \includegraphics[width=0.1035\textwidth]{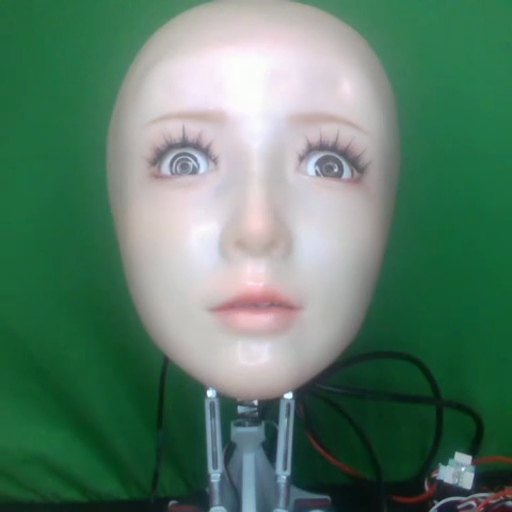} & \includegraphics[width=0.1035\textwidth]{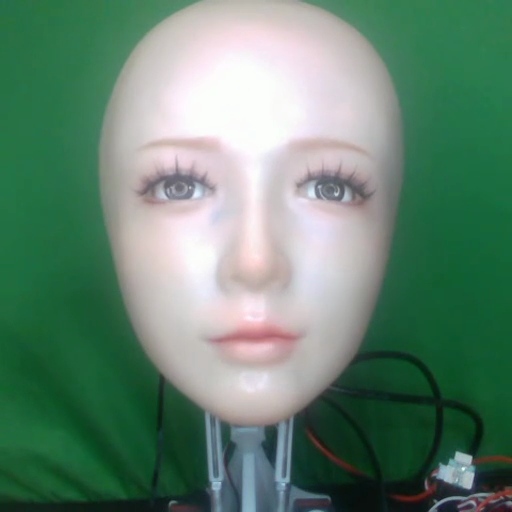} & \includegraphics[width=0.1035\textwidth]{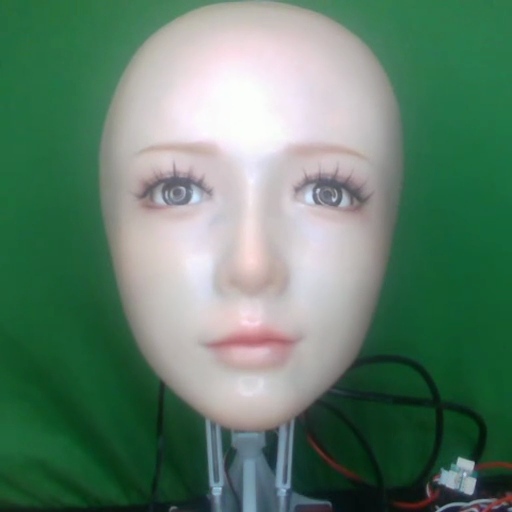} & \includegraphics[width=0.1035\textwidth]{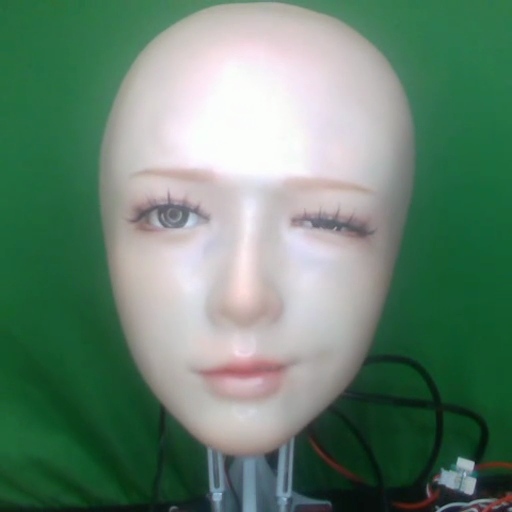} & \includegraphics[width=0.1035\textwidth]{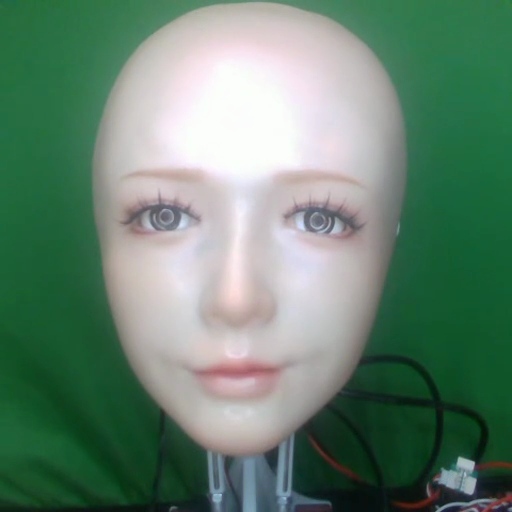} \\
\end{tabular}\endgroup
  }
  \caption{\textbf{Temporal expression examples on the ROSE face robot.} Each row shows nine real video frames of one expression sequence, arranged into three stages with the visible action described above the frames.}
  \label{fig:app:qualitative-dynamics}
  \vspace{-2.0em}
\end{figure*}

\bibliographystyle{IEEEtran}
\bibliography{main}
\end{document}